\documentclass[
]{ceurart}

\usepackage{listings}
\begin{document}

\copyrightyear{2024}
\copyrightclause{Copyright for this paper by its authors.
  Use permitted under Creative Commons License Attribution 4.0
  International (CC BY 4.0).}

\conference{CLEF 2024: Conference and Labs of the Evaluation Forum, September 09–12, 2024, Grenoble, France}

\title{SmurfCat at PAN 2024 TextDetox: Alignment of Multilingual Transformers for Text Detoxification}

\title[mode=sub]{Notebook for the PAN Lab at CLEF 2024}

\author[1]{Elisei Rykov}[%
email=Elisei.Rykov@skoltech.ru,
]
\cormark[1]
\author[2]{Konstantin Zaytsev}[%
email=kzaytsev@hse.ru,
]
\cormark[1]
\author[1]{Ivan Anisimov}[%
email=Ivan.Anisimov@skoltech.ru,
]
\author[1]{Alexandr Voronin}[%
email=Alexandr.Voronin@skoltech.ru,
]
\address[1]{Skolkovo Institute of Science and Technology, Russia}
\address[2]{HSE University, Russia}

\cortext[1]{These authors contributed equally.}

\begin{abstract}
This paper presents a solution for the \textit{Multilingual Text Detoxification} task in the PAN-2024 competition of the \textit{SmurfCat} team. Using data augmentation through machine translation and a special filtering procedure, we collected an additional multilingual parallel dataset for text detoxification. Using the obtained data, we fine-tuned several multilingual sequence-to-sequence models, such as \texttt{mT0} and \texttt{Aya}, on a text detoxification task. We applied the ORPO alignment technique to the final model. Our final model has only 3.7 billion parameters and achieves state-of-the-art results for the Ukrainian language and near state-of-the-art results for other languages. In the competition, our team achieved first place in the automated evaluation with a score of 0.52 and second place in the final human evaluation with a score of 0.74.\end{abstract}

\begin{keywords}
    PAN 2024 \sep
    Multilingual Text Detoxification \sep
    mT0 \sep
    ORPO
\end{keywords}

\maketitle

\section{Introduction}
Multilingual text detoxification is a challenging subtask within text style transfer. The most difficult part is the adaptation of such a system to low-resource languages. The concept of PAN-2024 Multilingual Text Detoxification Task \cite{bevendorff:2024, dementieva2024overview} is to develop a multilingual text detoxification system for 9 languages: Amharic, Arabic, German, Spanish, Hindi, Chinese, Russian, Ukrainian and English.

This paper describes the solution of the \textit{SmurfCat} team, which achieved first place with an average score of 0.52 in the automatic evaluation and second place with a score of 0.74 in the manual human evaluation. Our solution is based on the \texttt{mT0} model family \cite{muennighoff-etal-2023-crosslingual}, which has powerful multilingual capabilities. We fine-tuned all our selected models to each language of the competition, and applied various data augmentation techniques. To improve detoxification, we performed hypothesis filtering using the diverse beam search algorithm \cite{diverse}. Finally, we applied ORPO \cite{hong2024orpo} alignment to enforce model predictions. Our 3.7-billion-parameter language model demonstrates state-of-the-art results for Ukrainian and near state-of-the-art results for other languages. We published the final best-performing model on the HuggingFace Hub\footnote{\url{https://hf.co/s-nlp/mt0-xl-detox-orpo}}. You can also find the training scripts and the extended data on GitHub\footnote{\url{https://github.com/s-nlp/multilingual-transformer-detoxification}}.

The rest of the paper is organized as follows: Section \ref{sec:data} discusses data augmentation strategies, Section \ref{sec:method} describes our final solution, and Section \ref{sec:results} presents the results and discussion. 

\begin{figure}
    \centering
\resizebox{14cm}{!}{
    \includegraphics{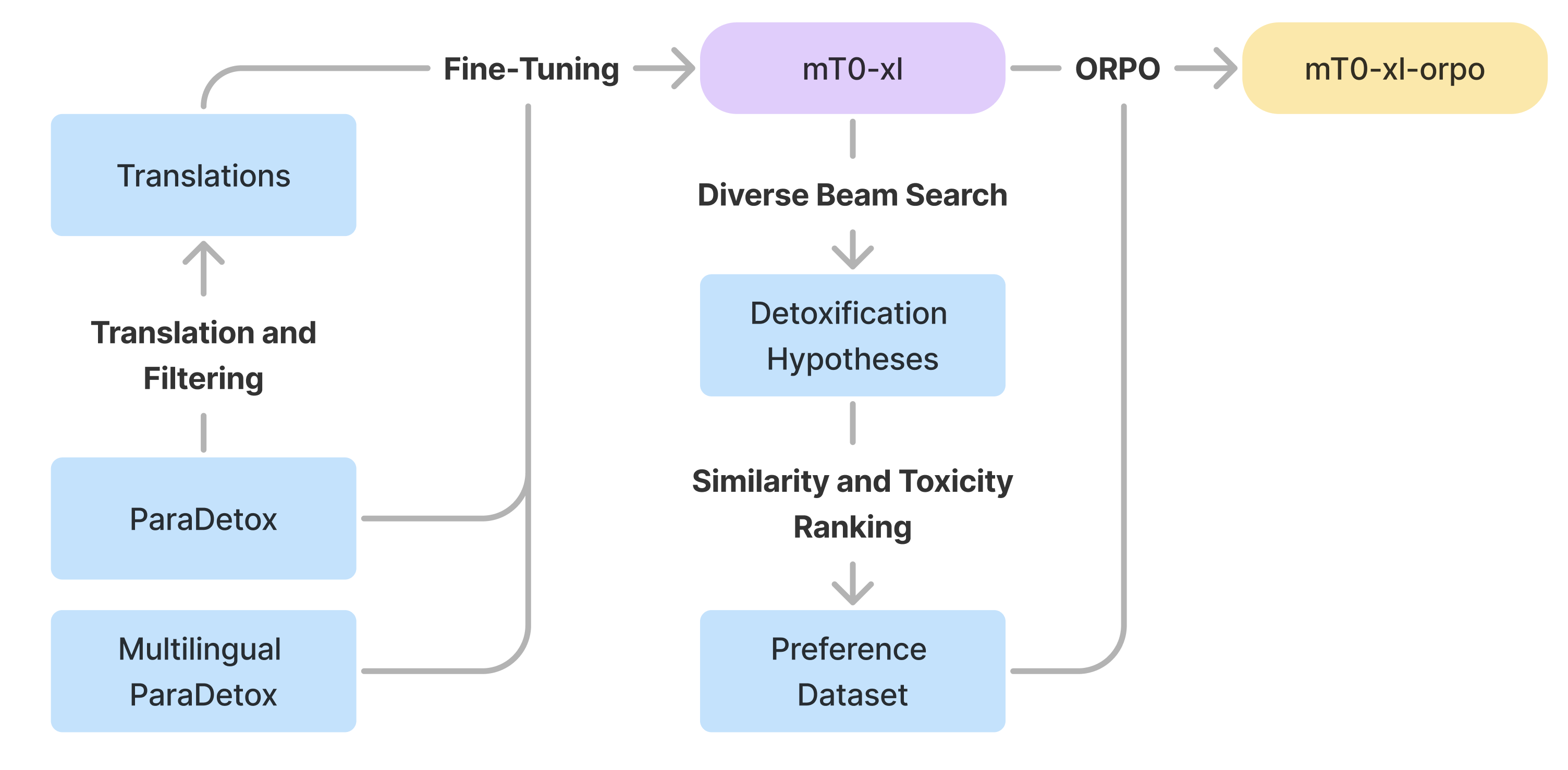}
}
    \caption{An overview of our approach. We used different datasets, fine-tuned the whole \texttt{mT0-XL} model and finally performed the ORPO alignment step.}
    \label{fig:overview}
\end{figure}

\section{Data}\label{sec:data}
Initially, there were not many parallel datasets for the multilingual detoxification task. More precisely, primarily only the Russian\footnote{\url{https://huggingface.co/datasets/s-nlp/ru_paradetox}} and English\footnote{\url{https://huggingface.co/datasets/s-nlp/paradetox}} ParaDetox datasets were available, with 11 100 and 19 700 samples respectively. During the competition, the organizers published a small human-annotated Multilingual ParaDetox\footnote{\url{https://huggingface.co/datasets/textdetox/multilingual_paradetox}} for all languages, containing only 400 samples per language.

Nevertheless, we decided to augment the provided data by automatic translation from English to other languages. To translate the original English data, we used a GoogleTranslator model from deep\_translator\footnote{\url{https://pypi.org/project/deep-translator/}} Python package. We chose API over some of the more advanced machine translation models because of its speed and simplicity. Also, there are not as many translators for low-resource languages like Amharic. As a result, we obtained an additional 19 700 samples for each language. 

\begin{figure}[h]
\centering
\begin{minipage}{.5\textwidth}
  \centering
  \includegraphics[width=1\linewidth]{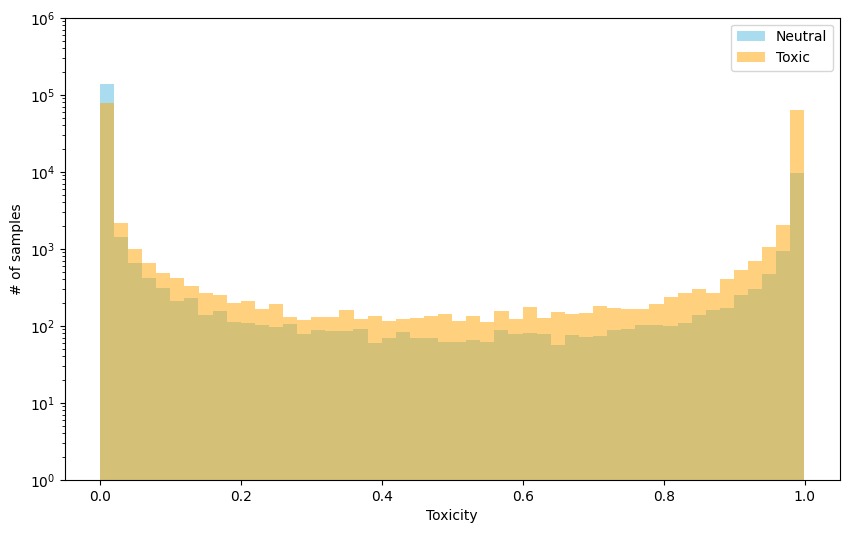}
  \captionof{figure}{Toxicity of translations}
  \label{fig:toxicity-distribution}
\end{minipage}%
\begin{minipage}{.5\textwidth}
  \centering
  \includegraphics[width=1\linewidth]{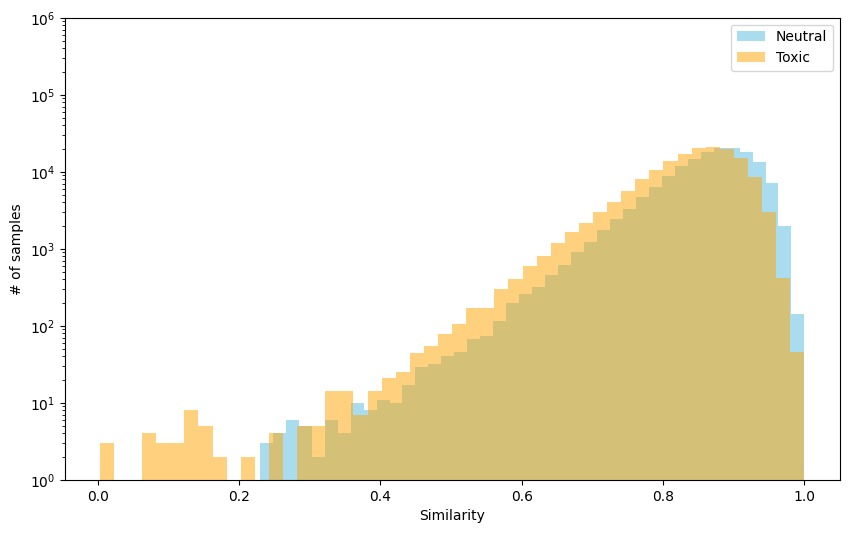}
  \captionof{figure}{Similarity of translations}
  \label{fig:similarity-distribution}
\end{minipage}
\end{figure}

Since translation is often imperfect, we decided to perform a specific post-processing procedure. In general, we checked the preservation of meaning after translation and the toxicity of the translated data. First, we used the LaBSE~\cite{labse} model to evaluate the similarity between translated pairs. Second, we applied XLM-R\footnote{\url{https://huggingface.co/textdetox/xlmr-large-toxicity-classifier}} toxicity classifier to check whether toxic sentences were still toxic after translation and vice versa.

A distribution for both of two measures is shown on Figures \ref{fig:toxicity-distribution}, \ref{fig:similarity-distribution}. For most samples, similarity between original and translated samples was high enough that many samples preserved their meaning. Regarding toxicity, many neutral sentences became toxic after translation, and many toxic sentences became neutral. For toxicity, we set a threshold parameter to 0.9 for toxic sentences and 0.1 for neutral sentences. The similarity threshold was set to 0.8 for all sentences. 

After all filtering steps, 40\,500 pairs of neutral and toxic sentences were obtained. A more precise statistic of how many samples remain after filtering is given in Table \ref{tab:data-stats}. According to the statistics, Amharic lost the most samples during filtering. 

\begin{table}[h]
\caption{Statistics of number of remaining samples after filtering.}
\begin{tabular}{@{}lllllllll@{}}
\toprule
\textbf{Language} & Amharic & Arabic & German & Spanish & Hindi & Russian & Ukrainian & Chinese   \\ \midrule
\textbf{\# of samples} & 1\,323 & 3\,190 & 7\,511 & 7\,555 & 4\,844 & 7\,458 & 5\,350 & 3\,274 \\ \bottomrule
\end{tabular}\label{tab:data-stats}
\end{table}

Our final dataset mixture is shown in Table \ref{tab:mixture}. In total, 74\,900 samples were used in the training process.

\begin{table}
\caption{Training dataset mixture.}
\begin{tabular}{@{}ll@{}}
\toprule
\textbf{Dataset}          & \textbf{\# of samples} \\ \midrule
En-ParaDetox                  & 19 700                 \\
Ru-ParaDetox                  & 11 100                 \\
Translations                  & 40 500                 \\
Multilingual ParaDetox & 3 600                  \\ \midrule
\textbf{Total}                & \textbf{74 900}        \\ \bottomrule
\end{tabular}
\label{tab:mixture}
\end{table}

\section{Method}\label{sec:method}
In this section, we describe our prior method, related to fine-tuning and optimization of Language Models on the text detoxification task.

\subsection{Supervised fine-tuning}
As a main approach, we choose fine-tuning of various multilingual LMs. As we suggest, the most promising models for the further fine-tuning were LMs from \texttt{mT0} family. It is a family of sequence-to-sequence Transformer models initialized from mT5 \cite{mt5}. We considered that sequence-to-sequence modeling would be more preferable in case of the text detoxification task. The \texttt{mT0} family was chosen because of its strong multilingual capabilities, so these models were adapted to each language of the competition. We also experimented with the novel \texttt{Aya-101} model \cite{aya-model}: a fine-tuned \texttt{mT5-xl} model on a multilingual instructions.

All models were tuned in an almost similar way. The learning rate was set to 1e-5, the global batch size to 8, and the weight decay to 0.01. The cosine scheduler was used for training. In total, 4 all models were trained during 4 epochs. All other training parameters were default according to HuggingFace \texttt{Seq2SeqTrainer}. The only difference is that for the \texttt{mT0-XL} we updated the weights of the whole model because our computing resources allowed it. In case of a larger model like \texttt{Aya-101} or \texttt{mT0-XXL}, only the LoRA adapter was trained. The setup of the LoRA adapter was as follows: r and lora alpha were set to 32, lora dropout parameter to 0.1, other parameters were default. The best model was selected according to the validation loss.

To enforce the in-context abilities of the models, we added a specific prefix to each toxic sentence depending on the language. As a result, we passed toxic sentences with special prefix prompt into the model during training.

\subsection{The Best Candidate Choice}
During inference, we generated 10 hypotheses and selected 5 most likely ones using diverse beam search. The number of beams was set to 10 with 5 beam groups, the diversity penalty was 2.5, the repetition penalty was 1.2. To select the best choice, we calculated a relevance metric using a product of similarity and toxicity scores. Similarity was calculated using \texttt{LaBSE} embeddings, and toxicity was measured using the \texttt{xlm-roberta-large} toxicity classifier. As relevance scores were calculated, we selected then the best candidate according to the highest score. 

\subsection{ORPO}
Once the models were fine-tuned, we decided to further tune the model for best performance using Odds Ratio Preference Optimization (ORPO) approach. This optimization does not need a reference model like it is in DPO \cite{rafailov2023direct}. Alignment was employed on the unseen test dataset.

As a preference dataset, we generated hypotheses using diverse beam search on the samples from the test set and annotated them using the relevance score described above. Only candidates with the highest relevance scores were selected as the chosen ones, and all others were selected as the rejected samples.

The final ORPO data set for alignment contained the prompt (toxic sentence), the rejected sample (negative candidate), and the selected sample (best candidate). Table \ref{tab:orpo-samples} shows a small sample of the dataset. Since the dataset was collected, we trained the model on the dataset using the same parameters used to train the other models. Since ORPO uses the beta parameter, it was set to 0.1. For the final submission, we used an aligned model with the algorithm described above to select the best candidate.
\begin{table}[h]
\caption{Samples for the ORPO tuning.}
\begin{tabular}{@{}p{5cm}p{5cm}p{5cm}}
\toprule
\textbf{Prompt}
& \textbf{Chosen}
& \textbf{Rejected}
\\ \midrule
Detoxify: She really can't be such a bitch, can she???
& She really can't be such a bad person, can she??? 
& really can't be such a bad person, can she???                            
\\ \midrule
Detoxify: hopefully turks are not dumb enough to hand it to him 
& Turkish people are not naive enough to hand it to him.
& hopefuly turks are not naive enough to hand it to him.
\\ \midrule
Detoxify: again , give me the name of the store or fuck off , liar . 
& again, give me the name of the store or go away        
& again, give me the name of the store
\\ \midrule
Detoxify: Nat is just a piece of shit, ignore him.   
& Nate is just not good, ignore him.   
& Nat is just a bad person, ignore him
\\ \bottomrule
\end{tabular}\label{tab:orpo-samples}
\end{table}


\section{Results}\label{sec:results}

\begin{table}[h]
\caption{The results of the automatic evaluation. The teams with the best scores were selected for the table. \textit{Joint} is given as the evaluation metric.}
\resizebox{\textwidth}{!}{
\begin{tabular}{@{}l|lllllllll|l@{}}
\toprule
\multicolumn{1}{c|}{\multirow{2}{*}{\textbf{Team}}} & \multicolumn{9}{c|}{\textbf{Language}}                                                                                                                                                                                          & \multirow{2}{*}{\textbf{Avg J}} \\ \cmidrule(lr){2-10}
\multicolumn{1}{c|}{}                                & \multicolumn{1}{c}{Amharic} & \multicolumn{1}{c}{Arabic} & \multicolumn{1}{c}{German} & \multicolumn{1}{c}{English} & \multicolumn{1}{c}{Spanish} & \multicolumn{1}{c}{Hindi} & \multicolumn{1}{c}{Russian} & \multicolumn{1}{c}{Ukrainian} & \multicolumn{1}{c|}{Chinese} &                               \\ \midrule
\textit{Our (mT0-XL-ORPO)} & \textit{\textbf{0.378}} & \textit{\textbf{0.626}} & \textit{\textbf{0.678}} & \textit{\textbf{0.602}} & \textit{\textbf{0.562}} & \textit{\textbf{0.355}} & \textit{\textbf{0.634}} & \textit{\textbf{0.692}} & \textit{\textbf{0.178}} & \textit{\textbf{0.523}} \\
\textit{Our (mT0-XL)} & \underline{\textit{0.374}} & \underline{\textit{0.617}} & \underline{\textit{0.669}} & \underline{\textit{0.593}} & \underline{\textit{0.555}} & \underline{\textit{0.352}} & \underline{\textit{0.628}} & \underline{\textit{0.686}} & \textit{0.165} & \underline{\textit{0.515}} \\
\textit{Our (mT0-XXL-LoRA)}  & \textit{0.361}   & \textit{0.594}  & \textit{0.639}   & \textit{0.591}   & \textit{0.548}   & \textit{0.345} & \textit{0.605} & \textit{0.660} & \textit{0.159} & \textit{0.500} \\
nikita.sushko  & 0.328 & 0.575 & 0.592 &   0.553     &  0.480    & 0.241 & 0.570     &   0.668   & \underline{0.176} & 0.465\\
VitalyProtasov  & 0.311 & 0.523 & 0.502 &   0.531    &  0.472    & 0.320 & 0.542     &   0.629   & 0.175 & 0.445\\
erehulka  & 0.287 & 0.536 &    0.575     &  0.543    &  0.497    & 0.185 & 0.529     &   0.602   & 0.160 & 0.435\\
\textit{Our (Aya-101-LoRA)} & \textit{0.301} & \textit{0.526} & \textit{0.530} & \textit{0.529} & \textit{0.475} & \textit{0.223} & \textit{0.541} & \textit{0.586} & \textit{0.108} & \textit{0.424} \\
ansafronov  & 0.270 & 0.456 &    0.362     &  0.506    &  0.319    & 0.133 & 0.507     &   0.328   & \textbf{0.178} & 0.340 \\ \bottomrule
\end{tabular}
}
\label{tab:results}
\end{table}

The final results of the automatic evaluation are shown in the Table \ref{tab:results}. The \texttt{mT0-XL} with ORPO alignment showed the best performance among all approaches from the leaderboard for all languages. Compared to \texttt{mT0-XL}, a model before ORPO alignment, ORPO slightly improved the performance of the model, increasing the average results by 0.01 points. Surprisingly, the larger models are not the best. For example, the \texttt{mT0-XXL} model with 13B parameters performed even worse than the \texttt{mT0-XL} model with only 3.7B parameters. \texttt{Aya-101}, an \texttt{mT5-XXL} model additionally tuned to instructional data for different languages, performed worse than other models. Since \texttt{Aya-101} and \texttt{mT0-XXL} performed even worse on \texttt{mt0-XL}, we did not perform an ORPO alignment step for these models. Considering other teams on the automatic evaluation, our checkpoints, mainly \texttt{mT0-XL-ORPO} and \texttt{mT0-XL}, are the two best performing approaches for all languages except the Chinese language.

The Table \ref{tab:evaluation} shows human evaluation results. Our detoxification model for Ukrainian achieved the highest human evaluation score by a wide margin, indicating that our approach is the state-of-the-art for this language. Overall, our best performing checkpoint is the top-2 approach according to the human evaluation by the averaged \textit{Joint} metric.

\begin{table}[h]
\caption{The results of the human evaluation. The teams with the best scores were selected for the table. \textit{Joint} is given as the evaluation metric.}
\resizebox{\textwidth}{!}{
\begin{tabular}{@{}l|lllllllll|l@{}}
\toprule
\multicolumn{1}{c|}{\multirow{2}{*}{\textbf{Team}}} & \multicolumn{9}{c|}{\textbf{Language}}                                                                                                                                                                                          & \multirow{2}{*}{\textbf{Avg J}} \\ \cmidrule(lr){2-10}
\multicolumn{1}{c|}{}                                & \multicolumn{1}{c}{Amharic} & \multicolumn{1}{c}{Arabic} & \multicolumn{1}{c}{German} & \multicolumn{1}{c}{English} & \multicolumn{1}{c}{Spanish} & \multicolumn{1}{c}{Hindi} & \multicolumn{1}{c}{Russian} & \multicolumn{1}{c}{Ukrainian} & \multicolumn{1}{c|}{Chinese} &                               \\ \midrule
Human Reference                                              & 0.85                  & 0.82                  & 0.71                  & 0.88                  & 0.79                  & 0.97                  & 0.80                  & 0.90                  & 0.93                   & 0.85                  \\ \midrule
SomethingAwful                                               & \underline{0.71}                  & 0.74                  & \textbf{0.89}                  & 0.86                  & \textbf{0.83}                  & \underline{0.86}                  & \textbf{0.84}                  & 0.69                  & 0.53                   & \textbf{0.77}                  \\
\textit{Our (mT0-XL-ORPO)}                                               & \underline{\textit{0.71}}                  & \underline{\textit{0.82}}                  & \textit{0.70}                  & \textit{0.83}                  & \textit{0.73}                  & \textit{0.68}                  & \textit{0.76}                  & \textit{\textbf{0.84}}                  & \textit{0.60}                   & \underline{\textit{0.74}}                  \\
VitalyProtasov                                               & 0.68                  & 0.79                  & 0.77                  & 0.69                  & \underline{0.81}                  & \textbf{0.87}                  & 0.73                  & 0.67                  & 0.49                   & 0.72                  \\
nikita.sushko                                               & 0.68                  & \textbf{0.89}                  & 0.79                  & 0.70                  & 0.62                  & 0.84                  & 0.74                  & 0.67                  & 0.47                   & 0.71                  \\
erehulka                                               & 0.69                  & 0.78                  & \underline{0.85}                  & 0.88                  & 0.71                  & 0.52                  & 0.65                  & 0.63                  & \underline{0.68}                   & 0.69                  \\
mkrisnai                                               & 0.49                  & 0.63                  & 0.70                  & \underline{0.89}                  & \textbf{0.83}                  & 0.73                  & \underline{0.78}                  & \underline{0.73}                  & 0.34                   & 0.68                  \\
d1n910                                               & 0.61                  & 0.44                  & 0.77                  & \textbf{0.91}                  & 0.77                  & 0.34                  & 0.71                  & 0.50                  & \textbf{0.84}                   & 0.65                  \\
ZhongyuLuo                                               & \textbf{0.72}                  & 0.49                  & 0.01                  & 0.73                  & 0.52                  & 0.49                  & 0.68                  & 0.42                  & 0.56                   & 0.51                  \\
\bottomrule
\end{tabular}
}\label{tab:evaluation}
\end{table}


\section{Conclusion}
In conclusion, our system demonstrated a strong pipeline for augmenting training data for low-resource languages and further fine-tuning a relatively small 3.7 billion parameter language model for the text detoxification task. Our future research may consider how to adapt text detoxification capabilities from high-resource languages to low-resource languages without translation, as machine translation for low-resource languages often shows low quality. A further direction for investigation is the interpretability of models, specifically the understanding of which tokens have been replaced by the model through the text detoxification process and the rationale behind this.


\bibliography{detox}

\end{document}